# Reference Resolution beyond Coreference: a Conceptual Frame and its Application

Andrei POPESCU-BELIS, Isabelle ROBBA and Gérard SABAH
Language and Cognition Group, LIMSI-CNRS
B.P. 133
Orsay, France, 91403
{popescu, robba, gs}@limsi.fr

**Abstract**

A model for reference use in communication is proposed, from a representationist point of view. Both the sender and the receiver of a message handle representations of their common environment, including *mental representations* of objects. Reference resolution by a computer is viewed as the construction of object representations using *referring expressions* from the discourse, whereas often only coreference links between such expressions are looked for. Differences between these two approaches are discussed. The model has been implemented with elementary rules, and tested on complex narrative texts (hundreds to thousands of referring expressions). The results support the mental representations paradigm.

## Introduction

Most of the natural language understanding methods have been originally developed on domain-specific examples, but more recently several methods have been applied to large corpora, as for instance morpho-syntactic tagging or word-sense disambiguation. These methods contribute only indirectly to text understanding, being far from building a conceptual representation of the processed discourse. Anaphora or pronoun resolution have also reached significant results on unrestricted texts. Coreference resolution is the next step on the way towards discourse understanding. The Message Understanding Conferences (MUC) propose since 1995 a coreference task: coreferring expressions are to be linked using appropriate mark-up.

Reference resolution goes further: it has to find out which object is referred to by an expression, thus gradually building a representation of the objects with their features and evolution. Coreference resolution is only part of this task, as coreference is only a relation between two expressions that refer to the same object.

A framework for reference use in human communication is introduced in Section 1, in order to give a coherent and general view of the phenomenon. Consequences for a resolution mechanism are then examined: data structures, operations, selectional constraints and activation. This approach is then compared to others in Section 2. Section 3 describes briefly the implementation of the model, the texts and the scoring methods. Results are given in Section 4, to corroborate the previous assertions and justify the model.

## 1 A general framework for reference use and resolution

### 1.1 Overview of the model

The communication situation is deliberately conceived here from a representationist point of view: the speaker (s) and the hearer (h) share the same world (W) considered as a set of objects with various characteristics or properties (Figure 1). Objects can be material or conceptual, or even belong to fictitious constructions. Each individual's perception of the world is different: $p_h(W) \neq p_s(W)$. Perception (p) as well as inferences (i) on perceptions using previous knowledge and beliefs provide each individual with a representation of the world, that is, $RW_s$ and $RW_h$, where $RW_x = i_x(p_x(W)) = i p_x(W)$. For computational reasons, it is useful to consider that only part of the world W plays a role in the communication act; this is called the topic T, and its representations are $RT_h$ and $RT_s$.

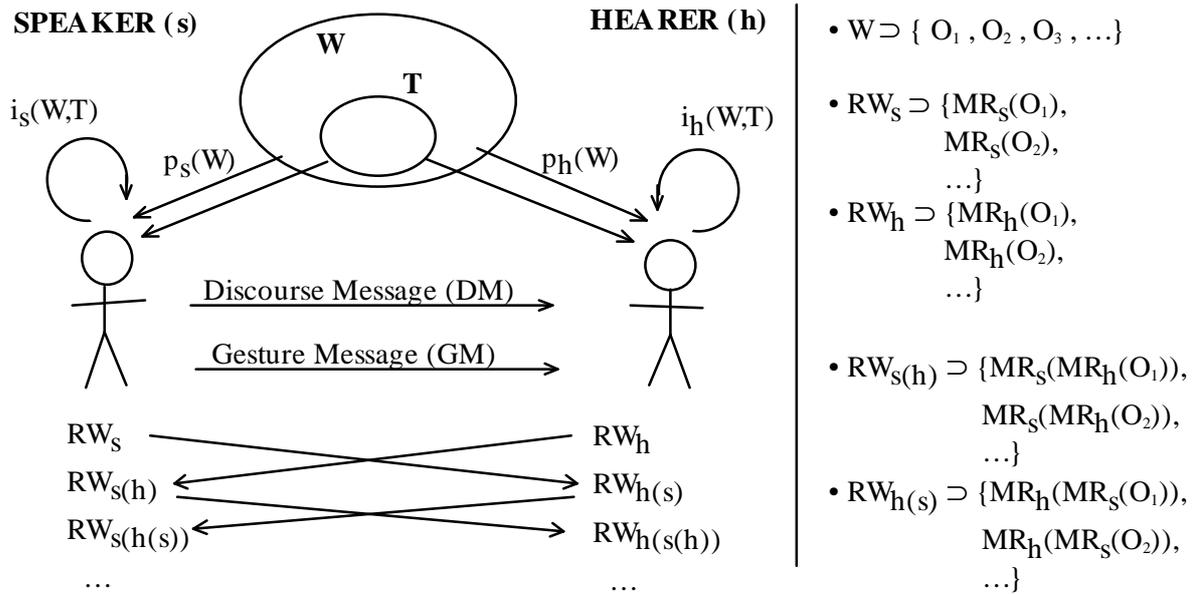

**Figure 1.** The proposed formal model for reference representation

The speaker produces a discourse message (DM) and a gesture message (GM). Both DM and GM contain referring expressions (RE), that is, chunks of discourse or gestures which are mapped to particular objects of RW. $RW_h$ and $RW_s$ each include a list of represented objects with their properties, called mental representations (MR). Understanding a message cannot be defined solely with respect to W, as there is no direct access to it. Instead, each individual builds a representation of the others' RW, using its own perceptions and inferences (ip). The speaker has his own $RW_s$ and also $RW_{s(h)} = ip_s(RW_h)$; the hearer has $RW_h$ and $RW_{h(s)} = ip_h(RW_s)$. This hierarchy, called specularity, is potentially infinite, as one may conceive $RW_{h(s(h))}$, $RW_{h(s(h(s)))}$, etc. (it could be tentatively asserted that when all the RW of all individuals become identical for a given assertion, the assertion becomes 'common knowledge').

A message has been understood if, for the current topic, $RT_{h(s)} = RT_s$, i.e., if the hearer's representation of the speaker's view of the world is accurate. This definition simplifies of course reality to make it fit into a computational model. For instance, from a rhetorical point of view, a communication succeeds if $RT_h$ changes according to the sender's will. Evolution in time isn't represented yet, so we do not index the various representations along the time axis.

In order to understand a message, the hearer has to find out which objects the referring expressions refer to: REs from the discourse, as well as deictic (pointing) ones. The hearer is able to use his own perception of W, namely $RW_h$, and his knowledge, to build mental representations of objects from the referring expressions.

### 1.2 Human-computer dialog vs. story understanding by a computer

We focus here on the problem of reference understanding by a computer program (c). Such a program has to build and manage, in theory, a $RW_c$ and a $RW_{c(s)}$, using information about the world, the message itself, and possibly a deictic set.

For a window manager application accepting natural language commands, the displayed graphic objects constitute the topic (T), i.e., the part of the world more specifically dealt with. The program's perception of T is totally accurate ($p_c(T) = T$); $p_c(T)$ is the most important and reliable source of information. Mouse pointing provides also direct deictic information. The difference between $RW_c$ and $RW_{c(s)}$ may account for the difference between the complete description of the displayed objects and their visible features.

For a story understanding program, the direct perception of the shared world W is strongly reduced, especially for fiction stories. Human readers in this case derive their knowledge only from the processed text. But knowledge about basic properties of W and about language conventions has still to be shared, otherwise no communication would be possible. For story processing, both $p_c(W)$ and the gesture message are extremely limited, so the program has to rely only on discourse information, thus building first $RW_{c(s)}$ and only afterwards $RW_c$, using supplementary knowledge about W. The gap between $RW_{c(s)}$ and $RW_c$ is due to the speaker's misuse of referring expressions, or to internal contradictions of the story. The system described below follows this second approach.

## 1.3 Data structures and operations

For minimal reference resolution, a program has to select the referring expressions (RE) of the received message and use them in order to build a list of mental representations of objects (MR). Each MR is a data structure having several attributes, depending on the program's capacities. Here is a basic set:
- MR.identificator – a number;
- MR.list-of-REs – the REs referring to the object;
- MR.semantic-information.text – a conceptual structure gathering the properties of the object, from the REs and from the sentences in which they appear;
- MR.semantic-information.dictionary – a conceptual structure gathering the properties of the object from the conceptual dictionary (concept lattice) of the system. These properties reflect a priori knowledge about the conceptual categories the MR belongs to;
- MR.relations – the relationship of the MR to other MRs, for instance: part-of or composed-of (these allow processing of plural MRs);
- MR.computer-object – a pointer on the object in case it belongs to a computer application (e.g., a window in a command dialog);
- MR.perceptual-information – an equivalent of the previous attribute, in case the program handles perceptual representations of objects.

In turn, the computational representation of a referring expression (RE) should have at least the following attributes:
- RE.identificator – a number;
- RE.position – uniquely identifies the RE's position in the text: number, paragraph, sentence, beginning and ending words;
- RE.syntactic-information – a parse tree of the RE, the RE's function, or, if available, a parse tree of the whole sentence where the RE appears;
- RE.semantic-information – a conceptual structure for the RE, or, if available, for the whole sentence.

Finally, there are operations on the MR set:
- creation: $RE_i \rightarrow MR_{new}$ – a new MR is created when an object is first referred to;
- attachment: $RE_i + MR_a \rightarrow MR_a$ – when a RE refers to an already represented object, the RE is attached to the MR and the MR's structure is updated;
- fusion: $MR_a + MR_b \rightarrow MR_{new}$ – at a given point, it may appear that two MRs were built for the same object, so they have to be merged. The symmetrical operation, i.e., splitting an MR which confusingly represents two objects, is far more difficult to do, as it has to reverse a lot of decisions;
- partition: $MR_a \rightarrow MR_a + MR_{new(1)} + MR_{new(2)} + \ldots$ ;
- grouping: $MR_a + MR_b \rightarrow MR_a + MR_b + MR_{new(a,b)}$;

The last two operations (partition/grouping) are symmetrical, and prove necessary in order to deal with collections of objects (plurals). For instance, from a collective RE as "the team" (and its MR) the program has to use built-in knowledge to create several MRs corresponding to the players, and correctly solve the new RE "the first player". Conversely, after construction of two MRs for "Miss X" and "Mrs. Y", an RE as "the two women" has to be attached to the MR which was built by grouping the previous MRs. In both cases, the MR.relation attribute has to be correctly filled-in with the type of relation between MRs.

If enough data is available, the system should build a conceptual structure for the MR (e.g., conceptual graphs), which should incrementally gather information from all referring expressions attached to the same MR. A lower-knowledge technique is to record for each MR a list of "characteristic REs" without any conceptual structures, and apply selectional constraints on it.

### 1.4 Selection heuristics

During the resolution process, each RE either triggers the creation of a new MR or is attached to an existing MR. The purpose of the selection heuristics is to answer whether the RE may be associated to a given MR, after examining compatibility between the RE and the other REs in the MR.list-of-REs. One of the simplest heuristics is:

- **(H1)** [$MR_a$ can be the referent of $RE_i$] *iff* [$RE_1$ being the first element of $MR_a$.list-of-REs, $RE_i$ and $RE_1$ can be coreferent]

This presupposes that the first RE referring to an object is typical, which isn't always true.

To take advantage of the MR paradigm, it may seem wiser to compare the current RE to all the REs in the MR.list-of-REs. This list includes also pronominal REs, which are actually meaningless for the compatibility test. Despite Ariel's (1990) claim that there is no clear-cut referential difference between pronouns and nominals, we will exclude pronouns in the implementation of our model. So, a second heuristic is:

- **(H2)** [$MR_a$ can be the referent of $RE_i$] *iff* [for all (non-pronominal) $RE_j$ in $MR_a$.list-of-REs, $RE_i$ and $RE_j$ can be coreferent]

This heuristic is in fact quite inefficient: first, it allows for little variation in the naming of a referent. Second, it neglects an important distinction in RE use, between identification and information (as described, for instance, by Appelt and Kronfeld (1987)). The sender may use a particular RE not only to identify the MR, but also to bring supplementary knowledge about it; thus, two REs conveying different pieces of knowledge may well be incompatible in the system's view. A more tolerant heuristic is thus:

- **(H3)** [$MR_a$ can be the referent of $RE_i$] *iff* [there exists a (non-pronominal) $RE_j$ in $MR_a$.list-of-REs so that $RE_i$ and $RE_j$ can be coreferent]

A more general heuristic subsumes both H2 ('all') and H3 ('one'):

- **(H4)** [$MR_a$ can be the referent of $RE_i$] *iff* [$RE_i$ and $RE_j$ can be coreferent for more than X% of the $RE_j$ in $MR_a$.list-of-REs]

When X varies from 0 to 100, this selection heuristic varies from H3 to H2 providing intermediate heuristics that can be tested (§4).

H3 seems in fact close to the co-reference paradigm, as it privileges links between individual REs, from which the MRs could even be built *a posteriori*, using the coreference chains. But here MRs are also characterized by an intrinsic activation factor, evolving along the text, which cannot be managed in the coreference paradigm.

### 1.5 Activation

The activation of an MR is computed according to salience factors (this technique is described for instance by Lappin and Leass (1994)). Our salience factors are: de-activation in time, re-activation by various types of RE, re-activation according to the function of the RE. Among the MRs which pass the selection, activation is used to decide whether the current RE is added to an MR (the most active) or if a new MR is created. Activation is thus a dynamic factor, which changes for each MR according to the position in the text and the previous reference resolution decisions.

## 2 Comparison with other works

Theoretical studies of discourse processing have long been advocating use of various representations for discourse referents. However, implementations of running systems have rather focused on anaphora or coreference. Our purpose here is to show how a simplified computational model of discourse reference can be implemented and give significant results for reference resolution; we showed previously (Popescu-Belis and Robba 1997) that it was also relevant for pronoun resolution.

### 2.1 High-level knowledge models

The idea of tracking discourse referents using 'files' for each of them has already been proposed by Kartunnen (1976). Evans (1985) and Recanati (1993) are both close to our proposals, however they neither give a computational implementation nor an evaluation on real texts. Sidner's work (1979) on focus led to salience factors and activations, but proved too demanding for an unrestricted use.

A more operational system using semantic representation of referents is for instance LaSIE (Gaizauskas et al. 1995), presented at MUC-6, which relies however a lot on task-dependent knowledge. The system doesn't seem to use activation cues. Another system (Luperfoy 1992) uses "discourse pegs" to model referents and was applied successfully to a man-machine dialogue task.

From a theoretical point of view, the model presented by Appelt and Kronfeld (1987) is in its background close to ours. Being further developed according to the speech acts theory, it relies however on models of intentions and beliefs of communicating agents which seem uneasy to implement for discourse understanding.

## 2.2 Robust, lower-level systems

Some of the robust approaches derive from anaphora resolution (e.g., Boguraev and Kennedy (1996)) because the antecedent / anaphoric links are a particular sort of coreference links, which disambiguate pronouns. Most of these systems however remain within the co-reference paradigm, as defined by the MUC-6 coreference task. Numerous low-level techniques have been developed, using generally pattern-matching between potentially coreferent strings (e.g., McCarthy and Lehnert 1995).

An interesting solution has been proposed by Lin (1995) using constraint solving to group REs into MRs. While this idea fits the MR paradigm, it doesn't work well incrementally, which makes use of activation impossible.

## 2.3 Advantages of the MR paradigm

Grouping REs into MRs brings decisive advantage even without conceptual knowledge. First, it suppresses an artificial ambiguity of coreference resolution: if RE1 and RE2 are already known as coreferent, coref(RE1, RE2), there is no conceptual difference between coref(RE3, RE1) and coref(RE3, RE2), so these two possibilities shouldn't be examined separately. Moreover, the system of coreference links makes it very time-consuming to find out whether REi and REj are coreferent, whereas MRs provide reusable storing of all the already acquired information.

Second, coreference links cannot represent multiple dependencies as needed by some objects which are collections of other objects. Coreference links simply mark identity of the referent for two REs: collections require typed links (part-of / composed-of) between several objects, as shown previously.

## 3 Application of the model

### 3.1 Reference resolution mechanism

We have particularized and implemented the theoretical model using algorithms in the style of Lappin and Leass (1994). We don't wish to overload this paper with technical details. The REs are solved one by one, either by attachment to an existent MR, or by creation of a new MR.

Selection rules are applied to the existing MRs to find out whether the current RE may or may not refer to the object represented by the MR. As our implementation deals with unrestricted texts, only very basic selection rules are used; there are two agreement rules (for gender and number) and a semantic rule (synonyms and hyperonyms are compatible). As no semantic network is available for French (e.g., WordNet), only very few synonyms are taken into account. Conceptual graphs are neither used, as our conceptual analyzer isn't robust enough for unrestricted noun phrases.

The working memory stores a fixed quota of the most active MRs, the others being archived and inaccessible for further resolution. From a cognitive point of view, this memory mimics the human incapacity to track too many story characters. Computationally, it reduces ambiguity for the attachment of REs, and increases the system's speed.

### 3.2 The texts

Two narrative texts have been chosen to test our system: a short story by Stendhal, *Vittoria Accoramboni* (VA) and the first chapter of a novel by Balzac, *Le PËre Goriot* (LPG) (Table 1). VA, available as plain text, underwent manual tagging of paragraphs, sentences and boundaries of all REs, then conversion to 'objects' of our programming environment (Smalltalk). Using

Vapillon's and al. (1997) LFG parser, an f-structure (parse tree) was added to each RE. Then the correct MRs were created using our user-friendly interface.

|  | VA | LPG.eq | LPG |
|---|---|---|---|
| Words | 2630 | 7405 | 28576 |
| REs | 638 | 686 | 3359 |
| MRs (key) | 372 | 216 | 480 |
| RE / MR | 1.72 | 3.18 | 7.00 |
| Nominal REs | 510 | 390 | 1864 |
| Pronoun REs | 102 | 262 | 1398 |
| Not parsed REs | 26 | 34 | 97 |

**Table 1.** Characteristics of the three texts.

LPG was already SGML-encoded with the REs and MRs, using Bruneseaux and Romary (1997) mark-up conventions. Only REs referring to the main characters of the first chapter were encoded: humans, places and objects. As a result, the ratio RE / MR is much greater than for VA. The text was converted to Smalltalk objects, f-structures were added to the REs, and MRs were automatically generated from the SGML tags. To make comparison with VA easier, a fragment of the LPG text was isolated (LPG.eq); it contains the same amount of REs as VA.

It should be noted that in both cases the LFG parser isn't robust enough to deliver proper f-structures for all noun phrases. The parser's total silence is ca. 4% and its ambiguity ca. 2.7 FS per RE. Despite such drawbacks (unreliable parser, lack of semantics), we kept working on complex narrative texts in order to study in depth the effects of elementary rules and parameters in situations where the coreference rate is high. Reference resolution is probably easier on technical documentation or articles, as referents receive more constant names.

### 3.3 Evaluation methods

The MRs produced by the reference resolution module (*response*) are compared to the correct solution (*key*) using an implementation of the algorithm described by Vilain and al. (1995), used also in the MUC evaluations. Although this algorithm was designed for coreference evaluation, it builds in fact each coreference chain, and compares the key and the response partition of the RE set in MR subsets — it follows thus the MR paradigm. The algorithm computes a *recall* error (number of coreference links missing in the response vs. the key) and a *precision* error (number of wrong coreference links, i.e. present in the response but absent from the key). The MUC scoring method isn't always meaningful. We have shown elsewhere (Popescu-Belis and Robba 1998) that it is too indulgent, and have proposed new algorithms which seem to us more relevant, named here 'core-MR' and 'exclusive-core-MR'.

### 4 Results and comments

The three heuristics H1, H2, H3 have been tested on our system, while keeping all other numeric parameters constant. The results Table 2 show that on average the heuristic H3 gives here the same results as H1, and is better than H2. As explained above, H2 is clearly too restrictive.

Different tests have been performed to analyze the system's results. If MR activation isn't used, the scores decrease dramatically, by ca. 50%. When using the H4 heuristic (variable average between H2 and H3) results aren't generally better than those of H3 (except for VA). Compatibility with only one RE of the MR seems thus a good heuristic.

|  | **H1** (first) | | **H2** (all) | | **H3** (one) | |
|---|---|---|---|---|---|---|
|  | R | P | R | P | R | P |
| MUC | .66 | .60 | .66 | .60 | .70 | .60 |
| Core | .52 | .44 | .52 | .44 | .56 | .39 |
| Ex-C | .62 | .73 | .63 | .73 | .60 | .69 |
| MUC | .72 | .76 | .66 | .70 | .72 | .76 |
| Core | .57 | .34 | .40 | .35 | .57 | .34 |
| Ex-C | .40 | .54 | .38 | .54 | .40 | 54 |
| MUC | .80 | .85 | .77 | .83 | .80 | .85 |
| Core | .38 | .40 | .34 | .42 | .38 | .40 |
| Ex-C | .29 | .48 | .28 | .48 | .29 | .48 |

**Table 2**. Success scores for selection heuristics (for VA, LPG.eq, LPG)

This is confirmed when applying the selection constraints on a limited subset of MR.list-of-REs. The worst results are obtained when this set fails to gather the shortest nonpronominal REs of an MR, which shows that these shortest strings (one or several) constitute a sort of 'standard name' for the referent, which suffices to solve the other references. The good score of H1 tends also to confirm this view.

An optimization algorithm based on gradient descent has been implemented to tune the activation parameters of the system. Not surprisingly, sometimes the local optimum has no cognitive relevance, as there is no searching heuristic other than recall+precision decrease. A local optimum obtained on one text still leads to good (but not optimal) scores on the other texts. Trained on VA, optimization led to a cumulated 4.3% improvement (precision + recall), and +2.5% on LPG.eq, or in another trial to +5.9%.

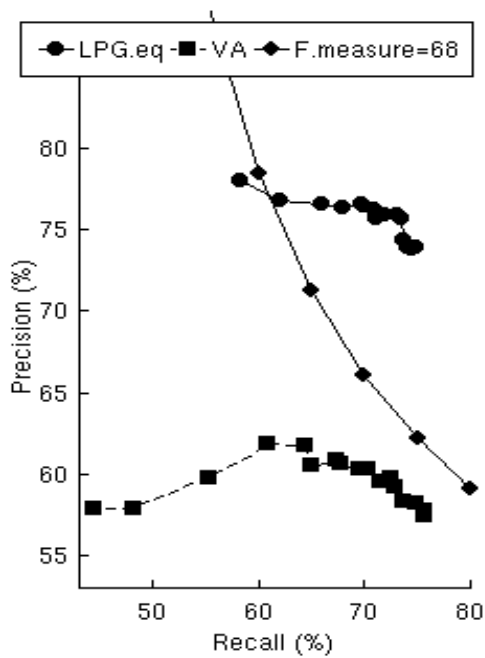

**Figure 2.** Influence of memory size on recall and precision (between 2, left, and 60, right)

Finally, the limited size buffer storing the MRs, a cognitively inspired feature, was studied. Variations of the system's performance according to the size of this 'working memory' show that it has an optimal size, around 20 MRs (Figure 2). A smaller memory increases recall errors, as important MRs aren't remembered. A larger memory leads to more erroneous attachments (precision errors) because the number of MRs available for attachment overpasses the selection rules' selectiveness.

## Conclusion

A theoretical model for reference resolution has been presented, as well as an implementation based on the model, which uses only elementary knowledge, available for unrestricted texts. The model shows altogether greater conceptual accuracy and higher cognitive relevance. Further technical work will seek a better use of the syntactic information; semantic knowledge will be derived in a first approach from a synonym dictionary, awaiting the development of a significant set of canonical conceptual graphs.

Further conceptual work, besides study of complex plurals, will concern integration of time to mental representations, as well as point of view information.

## Acknowledgments

The authors are grateful to F. Bruneseaux and L. Romary for the LPG text, to A. Reboul for discussions on the model, and to one of the anonymous reviewers for very significant comments. This work is part of a project supported by the GIS–Sciences de la Cognition.

## References

Appelt D. and Kronfeld A. (1987) *A Computational Model of Referring*, IJCAI '87, Milan, volume 2/2, pp. 640-647.

Ariel M. (1990) Accessing noun-phrase antecedents, Routledge, London.

Bruneseaux F. and Romary L. (1997) *Codage des références et coréférences dans les dialogues homme-machine,* ACH-ALLC '97, Kingston, Ontario, Canada.

Evans G. (1985) *The Varieties of Reference*, Oxford University Press, Oxford, UK.

Gaizauskas R., Wakao T., Humphreys K., Cunningham H. and Wilks Y. (1995) *University of Sheffield: Description of the LaSIE System as used for MUC-6*, MUC-6, pp. 207-220.

Kennedy C. and Boguraev B. (1996) *Anaphora in a Wider Context: Tracking Discourse*

*Referents,* ECAI 96, Budapest, Hungary, pp. 582-586.

Karttunen L. (1976) *Discourse referents.* In 'Syntax and Semantics 7: Notes from the Linguistic Underground', J. D. McCawley, ed., Academic Press, New York, pp. 363-385.

Lappin S. and Leass H. J. (1994) *An Algorithm for Pronominal Anaphora Resolution*, Computational Linguistics, 20/4, pp. 535-561.

Lin D. (1995) *University of Manitoba: Description of the PIE System Used for MUC-6*, MUC-6, pp. 113-126.

Luperfoy S. (1992) *The Representation of Multimodal User Interface Dialogues Using Discourse Pegs,* 30th Annual Meeting of the ACL, University of Delaware, Newark, Delaware, pp. 22-31.

McCarthy J. F. and Lehnert W. G. (1995) *Using Decision Trees for Coreference Resolution*, IJCAI '95, Montréal, Canada, pp. 1050-1055.

Popescu-Belis A. and Robba I. (1997) *Cooperation between Pronoun and Reference Resolution for Unrestricted Texts*, ACL'97 Workshop on Operational Factors in Practical, Robust Anaphora Resolution for Unrestricted Texts, Madrid, Spain, pp. 94-99.

Popescu-Belis A. and Robba I. (1998) *Three New Methods for Evaluating Reference Resolution*, LREC'98 Workshop on Linguistic Coreference, Granada, Spain.

Recanati F. (1993) *Direct Reference: from Language to Thought*, Basil Blackwell, Oxford, UK.

Sidner C. L. (1979) *Towards a computational theory of definite anaphora comprehension in English discourse*, Doctoral Dissertation, Artificial Intelligence Laboratory, Massachusetts Institute of Technology, Technical Report 537.

Vapillon J., Briffault X., Sabah G. and Chibout K. (1997) *An Object-Oriented Linguistic Engineering Environment using LFG (Lexical Functional Grammar) and CG (Conceptual Graphs),* ACL'97 Workshop on Computational Environments for Grammar Development and Linguistic Engineering, Madrid, Spain.

Vilain M., Burger J., Aberdeen J., Connolly D. and Hirshman L. (1995) *A Model-Theoretic Coreference Scoring Scheme,* 6th Message Understanding Conference, Columbia, Maryland.